\documentclass{article}
\usepackage{spconf,amsmath,graphicx}
\usepackage{multirow}
\usepackage{amsfonts}
\usepackage{url}

\title{PART REPRESENTATION LEARNING WITH TEACHER-STUDENT DECODER FOR OCCLUDED PERSON RE-IDENTIFICATION}
\name{Shang Gao$^{1,3}$ \qquad Chenyang Yu$^{1}$ \qquad Pingping Zhang$^{2,*}$\thanks{*Corresponding author (zhpp@dlut.edu.cn).} \qquad Huchuan Lu$^{1,2,3}$}
\address{$^{1}$ School of Information and Communication Engineering, Dalian University of Technology \\
      $^{2}$ School of Future Technology, School of Artificial Intelligence, Dalian University of Technology\\
      $^{3}$ NingBo Institute of Dalian University of Technology}
\begin{document}
%
\maketitle
\begin{abstract}
Occluded person re-identification (ReID) is a very challenging task due to the occlusion disturbance and incomplete target information.
Leveraging external cues such as human pose or parsing to locate and align part features has been proven to be very effective in occluded person ReID.
Meanwhile, recent Transformer structures have a strong ability of long-range modeling.
Considering the above facts, we propose a Teacher-Student Decoder (TSD) framework for occluded person ReID, which utilizes the Transformer decoder with the help of human parsing.
More specifically, our proposed TSD consists of a Parsing-aware Teacher Decoder (PTD) and a Standard Student Decoder (SSD).
PTD employs human parsing cues to restrict Transformer's attention and imparts this information to SSD through feature distillation.
Thereby, SSD can learn from PTD to aggregate information of body parts automatically.
Moreover, a mask generator is designed to provide discriminative regions for better ReID.
In addition, existing occluded person ReID benchmarks utilize occluded samples as queries, which will amplify the role of alleviating occlusion interference and underestimate the impact of the feature absence issue.
%
Contrastively, we propose a new benchmark with non-occluded queries, serving as a complement to the existing benchmark.
Extensive experiments demonstrate that our proposed method is superior and the new benchmark is essential.
The source codes are available at \url{https://github.com/hh23333/TSD}.
\end{abstract}
\begin{keywords}
Occluded Person Re-identification, Vision Transformer, Part Representation, Feature Distillation.
\end{keywords}
\begin{figure}[t]
\includegraphics[width=0.48\textwidth]{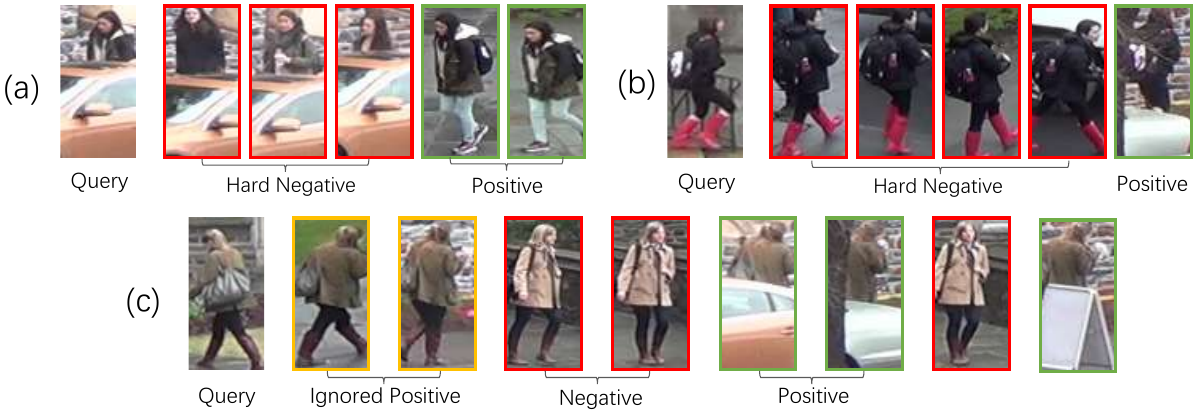}
\vspace{-8mm}
\caption{(a) An example with the interference of similar occlusions (from the Occluded-Duke benchmark~\cite{miao2019pose}). (b) An example with similar hard negative samples and the missing part in the positive sample (from our proposed benchmark). (c) The ranking setting in our benchmark, where holistic positive samples are ignored when calculating evaluation metrics.}
\vspace{-6mm}
\label{fig:intro}
\end{figure}
\section{INTRODUCTION}
\label{sec:intro}
Person re-identification (ReID) is the task of retrieving the query pedestrian across multiple non-overlapping cameras.
In recent years, significant improvements have been achieved with the advancement of deep learning~\cite{dosovitskiy2010image,he2016deep,wang2020deep}.
However, the accuracy of person ReID under occlusion is still unsatisfactory, which is a common situation in practical scenarios.

The difficulty of occluded person ReID is primarily due to the occlusion disturbance and misalignment caused by incomplete target information.
To address these issues, many previous methods~\cite{gao2020pose,somers2023body,he2019foreground,jia2022learning,wang2022pose} leverage external cues such as human pose or parsing to locate the human body parts and align the visible ones.
Recently, inspired by the strong generality in various vision tasks~\cite{carion2020end,ci2023unihcp,cheng2022masked}, some methods~\cite{zhang2021hat,jia2022learning,wang2022pose} propose to adopt Transformer structures to extract and disentangle deep features for occluded person ReID.
For example, Wang \emph{et al.}~\cite{wang2022pose} integrate patch features and key-point information with the Transformer decoder to enhance local patch features.
%
%
Jia \emph{et al.}~\cite{jia2023semi} propose a semi-attention partition method,
aiming to comply consistency with human parsing while keeping resistance against noisy supervision.
However, they may suffer from the part misalignment problem.
%
In this paper, a Teacher-Student Decoder (TSD) framework is proposed to utilize the Transformer decoder and focus on the corresponding body parts with the help of human parsing.
\begin{figure*}
\centering
\includegraphics[width=0.9\textwidth]{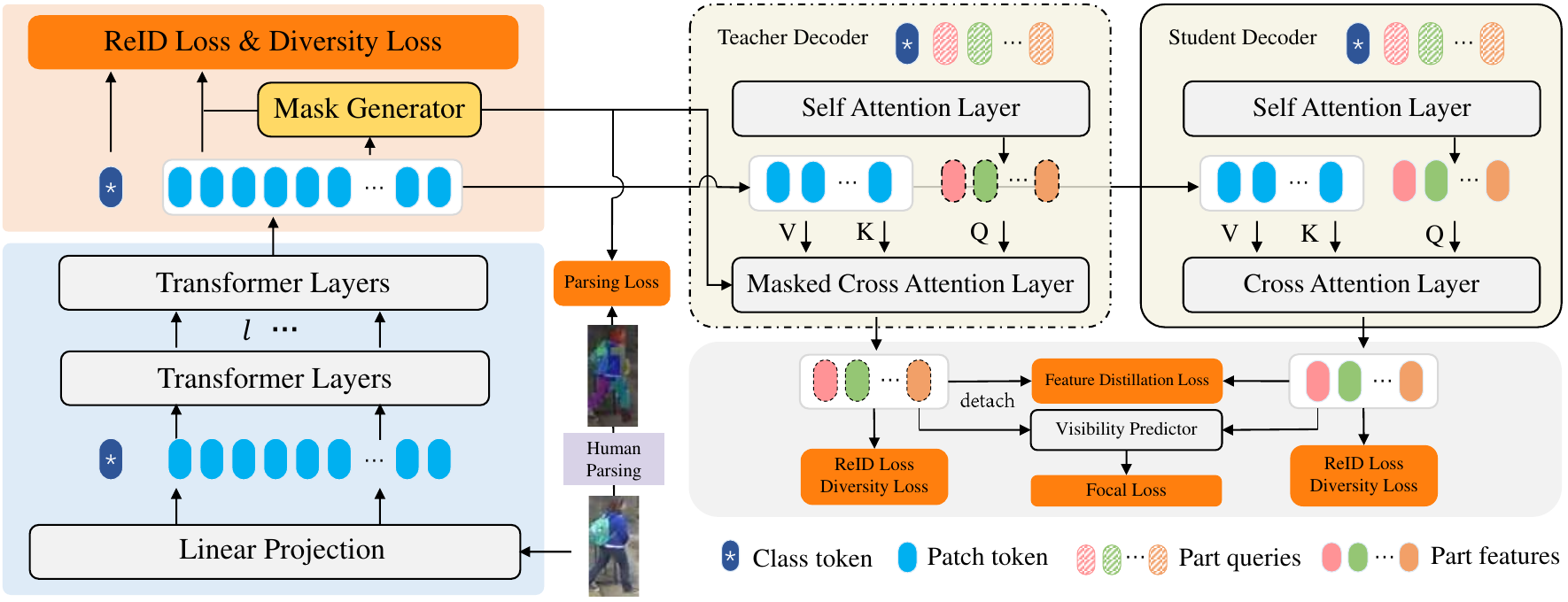}
\vspace{-4mm}
\caption{Overview of our proposed approach.}
\label{framwork}
\vspace{-4mm}
\end{figure*}
Specifically, our TSD consists of a Parsing-aware Teacher Decoder (PTD) and a Standard Student Decoder (SSD).
PTD restricts the attentions to specific body parts by human parsing cues.
Then, a feature distillation is employed on the outputs of PTD and SSD to ensure them closer to each other.
Thereby, SSD can learn from PTD to aggregate information from corresponding body parts automatically.

Meanwhile, existing occluded person ReID datasets only use occluded samples as queries.
As shown in Figure 1(a), the main concern is to eliminate the interference of similar occlusions. 
Therefore, it may underestimate the impact of the feature absence problem on evaluation metric.
To solve these problems, we propose a new benchmark for occluded person ReID.
As depicted in Figure 1(c), the query is a holistic sample and the positive holistic samples are ignored in the ranking list.
As shown in Figure 1(b), the query and the hard negative samples have similar lower bodies, while the lower body of the positive sample is occluded.
In this case, merely alleviating the noise caused by occlusion is insufficient.
The misalignment problem caused by the missing local information also needs to be addressed, thereby it can be reflected in evaluation metric.
Therefore, our benchmark can serve as a supplement to existing evaluation settings, and reflect the ReID performance under occlusion more comprehensively.

In summary, the main contributions of this work are three-fold:
(1) A Teacher-Student Decoder (TSD) framework is proposed to incorporate the human parsing information into the Transformer for occluded person ReID.
(2) A new benchmark is introduced to better evaluate the performance of occluded person ReID methods.
(3) Extensive experiments on existing benchmarks and the proposed benchmark demonstrate the superiority of our method.
\vspace{-2mm}
\section{METHODOLOGY}
\vspace{-2mm}
Our proposed framework is illustrated in Figure~\ref{framwork}.
%
%
Our method adopts a pre-trained ViT~\cite{dosovitskiy2010image} as the encoder to extract feature representations from the input image $X$.
The global feature $F^{g}\in \mathbb{R}^D$ is the final output of the class token and $F^{pt}\in \mathbb{R}^{N\times D}$ is the final patch embedding.
Here, $N$ is the number of patches and $D$ is the feature dimension.
We elaborate on key components in the following sections.
%
%
%
\vspace{-2mm}
\subsection{Teacher-Student Decoder}
\label{sec:decoder}
Our proposed TSD consists of two key components, i.e., a Parsing-aware Teacher Decoder (PTD) and a Standard Student Decoder (SSD).
They have the same structure and share parameters.
The only difference is that the teacher decoder takes external binary masks $M\in R^{P\times N}$ as inputs and restricts the discriminative regions for cross-attention.
$P$ is the number of parts.
$M(p,i)=1$ if the $i$\text{-}th location belongs to the $p$\text{-}th part and $M(p,i)=0$ vice versa.

We define $P$ semantic queries $Q^{em}=[q_1, q_2, ..., q_{P}]$ by a set of learnable embedding, each of which corresponds to a body part.
Then they are concatenated with the class embedding and fed into the standard self-attention layer as:
\begin{equation}
[q^{g}, Q^{em}] = \text{MHSA}([F^{g}, Q^{em}]).
\end{equation}
The MHSA is the Multi-Head Self-Attention.
The reason for incorporating the class token is that it helps the part queries better aggregate the features of the target pedestrian by the following cross attention layer.
In addition, the SSD directly employs the standard cross-attention as:
\begin{equation}
X^s = Softmax(QK^T/\sqrt{D})V,
\end{equation}
where $Q = \varphi(Q^{em}) \in R^{P\times D}$ is the query embedding under linear projection $\varphi$.
$K=\phi(F^{pt})\in R^{N\times D}$  and $V=\psi(F^{pt})\in R^{N\times D}$ are the image features under the transformations $\phi$ and $\psi$, respectively.
Finally, the output of the student decoder can be obtained by:
\begin{equation}
F^{sd} = \text{FFN}(\text{LN}(X^s)),
\end{equation}
where FFN and LN denote the feed-forward network and layer normalization~\cite{ba2016layer}, respectively.
The PTD employs masked attentions via:
\begin{equation}
X^t_p = Softmax(H_p+Q_pK^T)V,
\end{equation}
where $X^t_p$ is the $p$\text{-}th part feature, $Q_p$ is the $p$\text{-}th query embedding and $H_p\in R^N$ is defined by:
\begin{equation}
H_p(i) = \left\{
\begin{aligned}
0 & , & \text{if} \  M(p,i)=1, \\
-\infty & , & \text{otherwise}.
\end{aligned}
\right.
\end{equation}
The output of the teacher decoder can be obtained by:
\begin{equation}
F^{td} = \text{FFN}(\text{LN}(X^t)),
\end{equation}
then the feature distillation loss is employed to ensure each student query focus on a specific body part:
\begin{equation}
L_{fd} = \frac{1}{P}\sum_{i}(1 - Sim(F^{sd}_i, F^{td}_i)),
\end{equation}
where $Sim(\cdot)$ denotes the cosine similarity.
In addition, $L_{fd}$ may lead the decoder to extract identical features.
Thus, a diversity loss is employed on the part features:
\begin{equation}
L_{div} = \frac{1}{P(P-1)} \sum_{i=1,i\ne j}^{P} \sum_{j=1}^{P} Sim(F^{td}_i, F^{td}_j).
\vspace{-2mm}
\end{equation}
\subsection{Mask Generation}
\vspace{-2mm}
\label{sec:mask}
The mask can be obtained by directly employing human parsing methods~\cite{li2020self}.
However, it suffers from domain gaps and the lack of end-to-end training leads to generic masks not suitable for ReID.
Therefore, we propose a mask generator with both a body part prediction objective and a ReID objective as in~\cite{somers2023body}.
The mask generator consists of a fully-connected layer with parameters $G\in R^{(P+1)\times D}$ followed by a softmax layer.
Thus, the part heatmaps $M$ can be obtained by:
\begin{equation}
M = Softmax(F^{pt}G^T).
\end{equation}
To generate the semantic mask as the input of PTD, we perform the $argmax$ operation on $M$.
\vspace{-2mm}
\subsection{Overall Training and Inference Procedure}
\label{sec:procedure}
\textbf{Training Procedure:}
To optimize the proposed framework, the overall loss is formulated as:
\begin{equation}
\nonumber
\small
L = L_{ce}(F^g) + L_{tri}(F^g) + L_{ce}(F_c^{sd}) + L^p_{tri}(F^{sd}) + L_{ce}(F_c^{td})
\end{equation}
\begin{equation}
\begin{split}
&+ L^p_{tri}(F^{td}) + L_m + L_{fd} + L_{div} + L_v,
\end{split}
\end{equation}
where $L_{ce}$ is the cross-entropy loss with the BNNeck trick~\cite{luo2019bag}.
$L_{tri}$ is the triplet loss~\cite{hermans2017defense}.
$F_c^{sd}$ and $F_c^{td}$ are produced by concatenating the $P$ part features along the channel dimension.
$L^p_{tri}$ is the part average triplet loss~\cite{somers2023body}.
$L_m$ is used to train the mask generation as in~\cite{somers2023body}:
\begin{equation}
L_m = L_{ce}(F_c^{part}) + L^p_{tri}(F^{part}) + L_{pa}(M)
\end{equation}
\begin{equation}
F_i^{part} = \frac{\sum_{l=0}^{N}M(l)F^{pt}(l)}{\sum_{l=0}^{N}M(l)}
\end{equation}
\begin{equation}
F_c^{part} = Concat(F_1^{part}, F_2^{part}, ..., F_P^{part}).
\end{equation}
Here, $L_{pa}$ is a cross-entropy loss with label smoothing.
%
$L_v$ is the focal loss~\cite{lin2017focal} to supervise the visibility $v_p$ by a binary classifier upon $F^{sd}$:
\begin{equation}
L_{v} = \left\{
\begin{aligned}
-\alpha (1-v_p)^{\gamma}\log(v_p) & , & \text{if} \  \hat{v}_p=1, \\
-(1-\alpha){v_p}^{\gamma}\log(1-v_p) & , & \text{otherwise},
\end{aligned}
\right.
\end{equation}
where $\hat{v}_p$ is the label of the $p$\text{-}th part visibility defined by the human parsing as in~\cite{somers2023body}, $\alpha$ and $\gamma$ are parameters to balance positive vs.\ negative and hard vs.\ easy samples, respectively.

\noindent\textbf{Inference Procedure:}
The mask generator and teacher decoder are not needed in the inference procedure.
The visibility-based part-to-part matching strategy~\cite{somers2023body, gao2020pose} is adopted to calculate the distance of the query $q$ and gallery sample $g$:
\begin{equation}
d = \frac{\sum_{i \in \{g, 1,...,P\}} (v_i^q \cdot v_i^g \cdot dist(F^q_i, F^g_i))}{\sum_{i \in \{g, 1,...,P\}}(v_i^q \cdot v_i^g)}.
\end{equation}
Here, $\{F_i | i \in 1,...,P\}$ is the part feature from the student decoder and $dist$ is the Euclidean distance.
Finally, the ranking list is achieved by sorting the distance.
\vspace{-2mm}
\section{New Benchmark}
\vspace{-2mm}
To thoroughly measure the ReID performance under occlusion,
we introduce the Re-Occluded-Duke dataset, which is a reorganized version of the Occluded-Duke dataset~\cite{miao2019pose}.
\vspace{-1mm}
\subsection{Properties of Re-Occluded-Duke}
We maintain the same training set as the Occluded-Duke dataset~\cite{miao2019pose}, which consists of 15,618 images from 702 identities.
Then, we merge the original query and gallery sets into a new gallery set, which contains 18,001 images from 1,110 identities.
For images from the identities that appear in the original query set, we manually annotate their occlusion status into non-pedestrian occlusion (NPO), non-target pedestrian occlusion (NTP) and holistic images.
%
%
%
From these annotated images, we select all holistic images as the candidate set and randomly sample up to 5 images per identity to the query set.
Finally, the query set contains 2538 holistic images from 517 identities.
\vspace{-1mm}
\subsection{New Evaluation Metrics}
\label{EM}
The Cumulative Matching Characteristic (CMC) curves and mean Average Precision (mAP) are common metrics for evaluating the performance of different person ReID methods.
In order to better evaluate occluded samples, we propose several new evaluation metrics.
Occ-CMC and Occ-mAP focus on the retrieval of occluded samples.
They ignore the correct matches of holistic samples in the rank list.
Similarly, NPO-CMC and NPO-mAP are metrics that consider non-pedestrian occlusion, while NTP-CMC and NPO-mAP are metrics that account for non-target pedestrian occlusion.
\vspace{-2mm}
\section{Experiments}
\label{sec:copyright}
\vspace{-2mm}
\subsection{Datasets and Evaluation Metrics}
We evaluate our model on the holistic DukeMTMC-reID dataset~\cite{zheng2017unlabeled} and the occluded Occluded-Duke dataset~\cite{miao2019pose} as well as our benchmark.
We report the Rank-1 and mAP for the former two datasets and report the metrics as described in Sec.~\ref{EM} for our benchmark.
\subsection{Implementation Details}
We adopt the ViT-B~\cite{dosovitskiy2010image} as our backbone.
All images are resized to $256 \times 128$.
The training images are augmented with random erasing~\cite{zhong2020random}, padding, and random cropping for all experiments.
The batch size is set to 64 with 4 images per ID.
The SGD optimizer is employed with a momentum of 0.9 and the weight decay of $1e^{-4}$.
The learning rate is initialized as 0.004 with a cosine learning rate decay.
We train the model for 120 epochs.
The number of parts $P$ is set to 8.
We adopt the same human parsing model as in~\cite{somers2023body}.
\begin{table}[]
\centering
\caption{Comparison with other state-of-the-art methods on Occluded-Duke and DukeMTMC-reID. $\ast$ indicates the backbone is with an overlapping stride setting. $\dag$ indicates it is reproduced by replacing the original backbone with ViT.}\label{sota}
\resizebox{.8\columnwidth}{!}{
\begin{tabular}{r|cc|cc}
\hline
\multicolumn{1}{l|}{\multirow{2}{*}{Method}} & \multicolumn{2}{c|}{Occluded-Duke} & \multicolumn{2}{c}{DukeMTMC-reID} \\ \cline{2-5}
\multicolumn{1}{l|}{} & Rank-1 & mAP  & Rank-1 & mAP  \\ \hline
ViT-B~\cite{dosovitskiy2010image}  & 61.5 & \multicolumn{1}{c|}{53.5} & 88.8 & 79.3 \\
TransReID~\cite{he2021transreid} & 64.2 & \multicolumn{1}{c|}{55.7} & 89.6 & 80.6 \\ \hline
BPBreID$\dag$~\cite{somers2023body}  & 66.0 & \multicolumn{1}{c|}{56.7} & 90.2    & 80.8    \\
PFD~\cite{wang2022pose}       & 67.7 & \multicolumn{1}{c|}{\textbf{60.1}} & \textbf{90.6} & \textbf{82.2} \\
FED~\cite{wang2022feature}       & 68.1 & \multicolumn{1}{c|}{56.4} & 89.4 & 78.0 \\
Ours      & \textbf{70.6} & \multicolumn{1}{c|}{57.3} & 90.2 & 81.7 \\ \hline
DPM*~\cite{tan2022dynamic}     & 71.4 & \multicolumn{1}{c|}{61.8} & 91.0 & 82.6 \\
SAP*~\cite{jia2023semi}      & 70.0 & \multicolumn{1}{c|}{62.2} & -    & -    \\
PFD*~\cite{wang2022pose}       & 69.5 & \multicolumn{1}{c|}{61.8} & \textbf{91.2} & \textbf{83.2} \\
Ours *    & \textbf{74.5} & \multicolumn{1}{c|}{\textbf{62.8}} & 90.8  & 82.8  \\ \hline
\end{tabular}
}
\vspace{-4mm}
\end{table}
\begin{table}[]
\centering
\caption{Comparison with other methods on our benchmark.}\label{sota_new}
\resizebox{.84\columnwidth}{!}{
\begin{tabular}{r|cc|cccc}
\hline
\multicolumn{1}{l|}{\multirow{2}{*}{Method}} & \multicolumn{2}{c|}{OCC} & \multicolumn{2}{c|}{NPO}          & \multicolumn{2}{c}{NTP}                              \\ \cline{2-7}
\multicolumn{1}{l|}{}                        & Rank-1       & mAP       & Rank-1 & \multicolumn{1}{c|}{mAP} & \multicolumn{1}{c}{Rank-1} & \multicolumn{1}{c}{mAP} \\ \hline
VIT-B~\cite{dosovitskiy2010image}  & 67.1 & 52.5 & 60.8 & \multicolumn{1}{c|}{51.1} & 60.1 & 51.4 \\
FED~\cite{wang2022feature}       & 63.9 & 47.4 & 57.6 & \multicolumn{1}{c|}{46.0} & 56.7 & 46.6   \\
BPBreID$\ast$~\cite{somers2023body}  & 67.8 & 54.1 & 61.5 & \multicolumn{1}{c|}{53.4} & 59.0 & 50.4 \\
DPM~\cite{tan2022dynamic}       & 69.2 & 53.5 & 62.0 & \multicolumn{1}{c|}{50.8} & 63.6 & 53.9 \\
PFD~\cite{wang2022pose}       & 70.9 & 55.7 & 64.8 & \multicolumn{1}{c|}{54.3} & \textbf{64.6} & \textbf{55.2} \\
Ours      & \textbf{71.4} & \textbf{58.7} & \textbf{68.0} & \multicolumn{1}{c|}{\textbf{61.5}} & 61.9 & 52.5 \\ \hline
SAP*~\cite{jia2023semi}      & 71.4 & 57.1 & 65.8 & \multicolumn{1}{c|}{55.4} & \textbf{65.4} & 56.6 \\
Ours *    & \textbf{73.2} & \textbf{61.7} & \textbf{68.8} & \multicolumn{1}{c|}{\textbf{62.7}} & 64.9 & \textbf{57.5} \\ \hline
\end{tabular}}
\vspace{-4mm}
\end{table}
\vspace{-2mm}
\subsection{Comparison with State-of-the-art Methods}
We compare our model with other state-of-the-art methods on Occluded-Duke and DukeMTMC-reID in Table~\ref{sota}.
All of them adopt Transformers as the backbone.
Our method outperforms all the competitors on Occluded-Duke dataset.
The main reason is that our TSD guides each query adaptively focus on a corresponding body part, thereby better disentangling pedestrian features.
Furthermore, our method achieves competitive results on the holistic dataset DukeMTMC-reID.

Table~\ref{sota_new} shows the result of our method and other methods on our new benchmark.
It can be observed that our method can achieves superior performance, especially on the NPO samples.
It is noteworthy that FED~\cite{wang2022feature} performs well on the Occluded-Duke dataset.
However, it is inferior on our new benchmark, even worse than the baseline method.
The main reason is that it only considers the noise introduced by occlusion, but ignores the misalignment issue.
%
\begin{table}[]
\centering
\caption{Ablation study for the main components on Occluded-Duke and our benchmark.}\label{ablation}
\resizebox{.88\columnwidth}{!}{
\begin{tabular}{r|cc|ccc}
\hline
\multirow{2}{*}{Method} & \multicolumn{2}{l|}{Occluded-Duke} & OCC & NPO & NTP \\ \cline{2-6}
                        & Rank-1  & mAP & mAP & mAP & mAP \\ \hline
Baseline & 61.5 & 53.5 & 52.3 & 50.2 & 50.8 \\
M1  & 59.9 & 51.5 &  49.0    & 47.4  & 48.4  \\
M2   & 65.1 & 54.3 & 54.6  & 57.8  & 48.6  \\
M3   & 68.3 & 54.9 & 54.5 & 56.8 & 48.2 \\
M4   & 70.6 & 57.3 & 58.7 & 61.5 & 52.5 \\ \hline
\end{tabular}
}
\vspace{-2mm}
\end{table}
\begin{figure}
\centering
\includegraphics[width=0.4\textwidth]{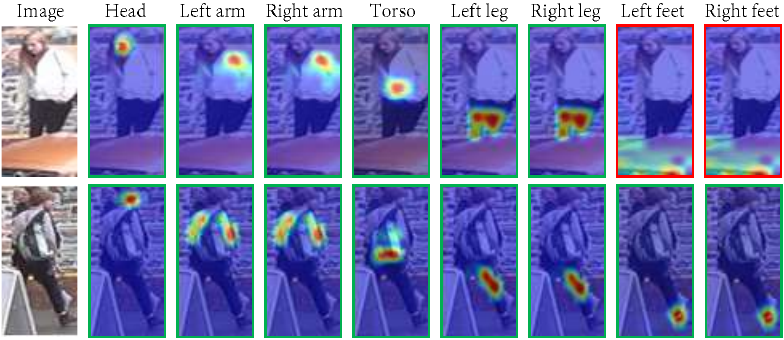}
\vspace{-4mm}
\caption{Visualization of attention maps. Green boxes indicate visible predictions, while red boxes for invisible ones.} \label{vis_attn}
\vspace{-4mm}
\end{figure}
\vspace{-2mm}
\subsection{Ablation Study}
In this section, we conduct ablation studies on the Occluded-Duke and our benchmark to analyze the effectiveness of each component.
The results are shown in Table~\ref{ablation}.
The baseline is with the standard vision Transformer~\cite{he2021transreid}.
M1 denotes the pure Transformer encoder-decoder architecture.
Its results are inferior to the baseline.
The possible reason is that the model can not adaptively decouple features in the absence of auxiliary supervision, and the newly initialized decoder may even bring a negative effect.
From M2, when the masked teacher is added, the performance is greatly improved.
This indicates that the masked teacher can effectively aggregate local features and transfer the information to guide the student decoder.
M3 denotes the diversity loss is further added to M2.
It can be observed that the diversity loss also brings performance improvement.
By replacing the fixed mask with a learnable one, M4 surpasses M3.
It shows our mask generator can effectively identify distinctive regions.
Furthermore, Figure~\ref{vis_attn} shows the attention maps towards each query of the decoder.
Our method can successfully focus on different body parts and precisely predict the visibility.
It clearly shows the effectiveness of extracting discriminative regions.
\vspace{-2mm}
\section{CONCLUSION}
\vspace{-2mm}
In this paper, a Teacher-Student Decoder (TSD) framework is proposed to incorporate the human parsing information into the Transformer structure for occluded person ReID. In addition, a new benchmark is introduced to better evaluate the ReID performance under occlusion. Extensive experiments demonstrate the superiority of our method.

\small
\textbf{Acknowledgments:} This work was supported in part by the National Key Research and Development Program of China (No. 2018AAA0102001) and National Natural Science Foundation of China (No. 62101092). We thank all reviewers for their comments.

\vfill\pagebreak



\bibliographystyle{IEEEbib}
\bibliography{strings,refs}

\end{document}